%% file: paper.subjective.wifs2020.tex
\documentclass[conference]{IEEEtran}
\usepackage[latin9]{inputenc}
\usepackage{times}
\usepackage{amsmath}
\usepackage{graphicx}
\usepackage{epsfig}
\usepackage{amssymb}
\usepackage{multirow}
\usepackage{booktabs}
\usepackage{subfig}

\usepackage[norule]{footmisc}

\usepackage{float} % Required for tables and figures in the multi-column environment - they need to be placed in specific locations with the [H] (e.g. 

\graphicspath{{figures/}}
\DeclareGraphicsExtensions{.eps,.png,.pdf,.jpg,.jpeg,.JPG}

\usepackage{url}
\begin{document}
\title{Deepfake detection: humans vs. machines}

% Yo
\author{\IEEEauthorblockN{Pavel Korshunov}
\IEEEauthorblockA{Idiap Research Institute \\
Martigny, Switzerland\\
Email: pavel.korshunov@idiap.ch}
\and
\IEEEauthorblockN{S\'{e}bastien Marcel}
\IEEEauthorblockA{Idiap Research Institute \\
Martigny, Switzerland\\
Email:  sebastien.marcel@idiap.ch}
}


%Telephone: (800) 555--1212, Fax: (888) 555--1212}


\maketitle

%INCLUDES COPYRIGHT NOTICE: one of three copyright notice should be included.
%Uncomment the appropriate line below, according to the authors %affiliation:
\begin{figure}[b]
\vspace{-0.3cm}
\parbox{\hsize}{\em
%information about the event:
%WIFS`2020, December, 6-9, 2020, New York, USA.
%copyright notice: one of four copyright notices below should be included. Choose the right one below according to the authors affiliation:
%XXX-X-XXXX-XXXX-X/XX/\$XX.00 \ \copyright 2017 European Union.
%XXX-X-XXXX-XXXX-X/XX/\$XX.00  \ \copyright 2017 Crown.
%U.S. Government work not protected by U.S. copyright.
%XXX-X-XXXX-XXXX-X/XX/\$XX.00 \ \copyright 2017 IEEE.
%XXX-X-XXXX-XXXX-X/XX/\$XX.00 \ \copyright 2020 IEEE.
}\end{figure}

\input{sections/abstract}


% \fi
\IEEEpeerreviewmaketitle

\input{sections/introduction}

\input{sections/dataset}

\input{sections/results}

\input{sections/conclusion}

\section*{Acknowledgements}
This work was funded by Hasler Foundation's VERIFAKE project and 
Swiss Center for Biometrics Research and Testing.



\bibliographystyle{IEEEtran}
\bibliography{references_pavel}
%

%
% <OR>
%
%
%manually copy in the resultant .bbl file
%\begin{thebibliography}{1}
%\end{thebibliography}

\end{document}

%% file: sections/abstract.tex
\begin{abstract}

Deepfake videos, where a person's face is automatically swapped with a face of someone else, are becoming easier to generate with more realistic results. In response to the threat such manipulations can pose to our trust in video evidence, several large datasets of deepfake videos and many methods to detect them were proposed recently. However, it is still unclear how realistic deepfake videos are for an average person and whether the algorithms are significantly better than humans at detecting them. In this paper, we present a subjective study conducted in a crowdsourcing-like scenario, which systematically evaluates how hard it is for humans to see if the video is deepfake or not. For the evaluation, we used 120 different videos (60 deepfakes and 60 originals) manually pre-selected from the Facebook deepfake database, which was provided in the Kaggle's Deepfake Detection Challenge 2020. For each video, a simple question: ``Is face of the person in the video real of fake?'' was answered on average by 19 na\"ive subjects. The results of the subjective evaluation were compared with the performance of two different state of the art deepfake detection methods, based on Xception and EfficientNets (B4 variant) neural networks, which were pre-trained on two other large public databases: the Google's subset from FaceForensics++ and the recent Celeb-DF dataset. The evaluation demonstrates that while the human perception is very different from the perception of a machine, both successfully but in different ways are fooled by deepfakes. Specifically, algorithms struggle to detect those deepfake videos, which human subjects found to be very easy to spot.
\end{abstract}

%% file: sections/introduction.tex
\section{Introduction}
\label{sec:intro}

Autoencoders and generative adversarial networks (GANs) significantly improved the quality and realism of the automated image generation and face swapping, leading to the deepfake phenomena. Many are starting to believe that the proverb `seeing is believing' is starting to loose its meaning when it comes to digital video\footnote{\url{https://edition.cnn.com/interactive/2019/01/business/pentagons-race-against-deepfakes/}}. The concern for the impact of the widespread deepfake videos on our trust in video recording is growing. This public unease prompted researchers to propose various datasets of deepfakes and methods to detect them. Some of the latest approaches demonstrate encouraging accuracy, especially, if they are trained and evaluated on the same datasets. 

%\begin{figure}[tb]
%\centering
%\includegraphics[width=0.8\columnwidth]{face-swapping-process}
%\caption{Process of generating deepfake videos.}
%\label{fig:process}
%\end{figure}

\begin{figure}[tbh]
\centering
\subfloat[By Google]{\includegraphics[width=0.19\textwidth]{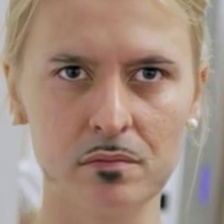}}
~
\subfloat[DeepfakeTIMIT]{\includegraphics[width=0.19\textwidth]{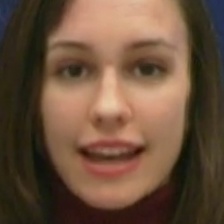}} \\
\subfloat[By Facebook]{\includegraphics[width=0.19\textwidth]{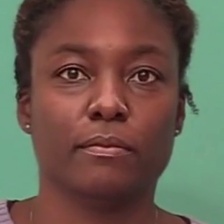}}
~
\subfloat[Celeb-DF]{\includegraphics[width=0.19\textwidth]{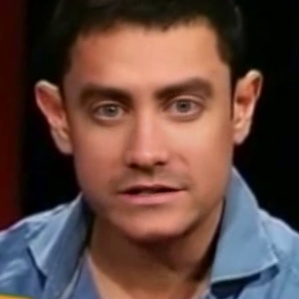}}
\\
  
\caption{Examples of deepfakes (faces cropped from videos) in different databases.}
\label{fig:diffrentdb}
\end{figure}

Many databases with deepfake videos were created to help develop and train deepfake detection methods. One of the first freely available database was based on VidTIMIT~\cite{Korshunov2019a}, followed by the FaceForeniscs database, which `deepfaked' $1'000$ Youtube videos~\cite{Verdoliva2018} and which later was extended with a larger set of high resolution videos provided by Google~\cite{Roessler2019}. Another recently proposed $5'000$ videos-large database of deepfakes generated from Youtube videos is Celeb-Df~\cite{Celeb_DF_cvpr20}. But the most extensive and the largest database to date with more than $100K$ videos (80\% of which are deepfakes) is the dataset from Facebook, which appeared in the recent Deepfake Detection Challenge hosted by Kaggle\footnote{\url{https://www.kaggle.com/c/deepfake-detection-challenge}}. Figure~\ref{fig:diffrentdb} shows examples of faces cropped from deepfake videos in various databases.

\begin{figure*}[tbh]
\centering

\subfloat{\includegraphics[width=0.195\textwidth]{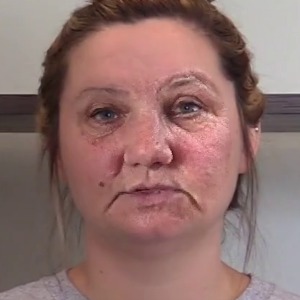}}
\subfloat{\includegraphics[width=0.195\textwidth]{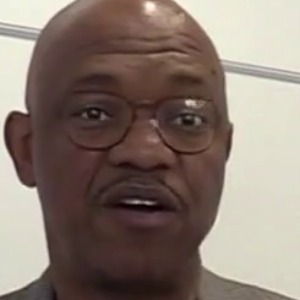}}
\subfloat{\includegraphics[width=0.195\textwidth]{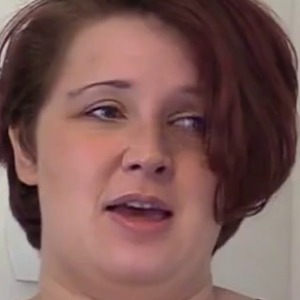}}
\subfloat{\includegraphics[width=0.195\textwidth]{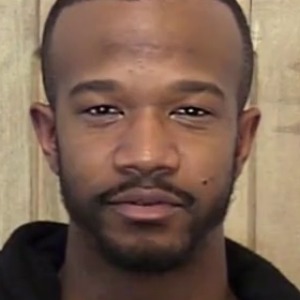}}
\subfloat{\includegraphics[width=0.195\textwidth]{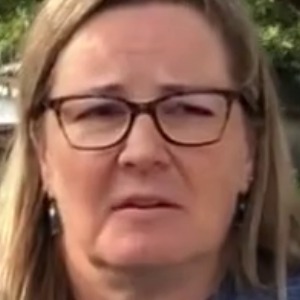}} \\
\subfloat[Very easy]{\includegraphics[width=0.195\textwidth]{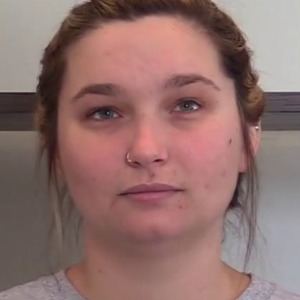}}
\subfloat[Easy]{\includegraphics[width=0.195\textwidth]{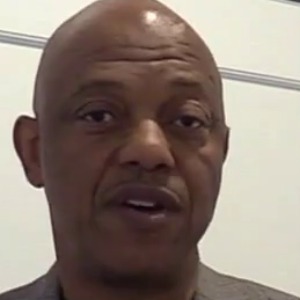}}
\subfloat[Moderate]{\includegraphics[width=0.195\textwidth]{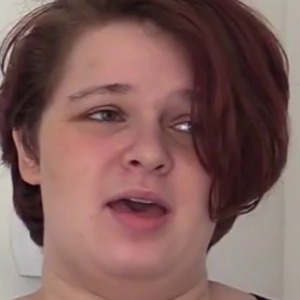}}
\subfloat[Difficult]{\includegraphics[width=0.195\textwidth]{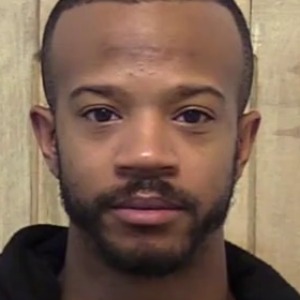}}
\subfloat[Very difficult]{\includegraphics[width=0.195\textwidth]{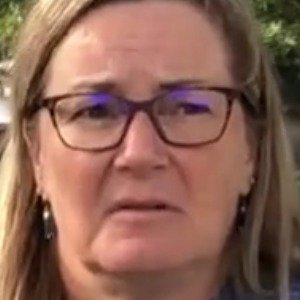}}
\\
  
\caption{Cropped faces from different categories of deepfake videos from Facebook database (top row) and the corresponding original versions (bottom row).}
\label{fig:examples}
\end{figure*}

These datasets were generated using either the popular open source code\footnote{\url{https://www.kaggle.com/c/deepfake-detection-challenge/discussion/121313}}, typically, deepfakes from Youtube videos, or the latest methods by Google and Facebook for creating deepfakes. The fact that even Google and Facebook, private companies who are typically very frugal with making large datasets publicly available, provided some of the most extensive datasets for research shows how important and challenging is the deepfake detection for the scientific and industrial communities. This abundance of deepfake video data allowed researchers to train and test detection approaches based on very deep neural networks, such as Xception ~\cite{Roessler2019}, capsules networks~\cite{Nguyen2019}, ResNet-50~\cite{Efros2020}, and EfficientNet~\cite{Delp2020} which were shown to outperform the methods based on shallow CNNs, facial physical characteristics~\cite{Li2018,Yang2019head,Agarwal2020detecting}, or distortion features~\cite{Zhang2017,Agarwal2017,korshunov2018deepfakes}. 

However, despite the public and media uneasiness with deepfake videos and the surge of automated methods for their detection, little is known about how `good` the deepfakes actually are at `fooling` human perception. Most of the public perception that deepfakes are realistic comes from personal experience of watching some video examples on Youtube, the alarming media reports, and the understanding that the deepfake generation technology will become more realistic in the nearest future. There is a lack of scientific studies on how  realistic the currently available deepfakes are and whether they can pose a threat to human perception of video. The only study~\cite{Roessler2019} that asked human subjects to evaluate $60$ images ($30$ were fake but the number of deepfakes was not reported) demonstrated that almost $80$\% of deepfake images were successfully recognized as fake. 

In this paper, we conducted a more comprehensive subjective evaluation  (of deepfake videos instead of images), using the web-based framework for crowdsourcing experiments QualityCrowd 2~\cite{QualityCrowd}. We want to understand how easily an average human observer can be spoofed by different types of deepfake videos. For that purpose, we selected $120$ videos ($60$ original and $60$ deepfakes) from  Facebook dataset\footnotemark[2], because it is the largest and one the most recent databases, and it has many different variants of deepfakes, ranging from the most obvious ones to those that look very realistic. We have defined five categories of deepfakes ($12$ of each) by judging them on how easy it is to spot their visual artifacts as `very easy', `easy', `moderate', `difficult', and `very difficult' (see Figure~\ref{fig:examples} for some examples). For each video, on average $20$ na\"ive subjects (including PhD students, senior scientists, and people in administration) had to answer if they think it is fake or not.

Understanding how well people recognize deepfake is important, but also is the understanding of how detection algorithms recognize them too. Policy decisions as well as people's perceptions are often based on the assumption that automated detection algorithms perceive videos in a way that is similar to humans\footnote{\url{https://www.forbes.com/sites/fernandezelizabeth/2019/11/30/ai-is-not-similar-to-human-intelligence-thinking-so-could-be-dangerous/}}, which can be even dangerous when it comes to such impactful technology as deepfake detection. 

Therefore, in this paper, we also assess how two state of the art algorithms, based on Xception model~\cite{Chollet2017} and EfficientNet variant B4~\cite{Delp2020}, both of which showed a great performance on several deepfake databases~\cite{Roessler2019}. pre-trained on two other large databases from Google~\cite{Roessler2019} (a subset of FaceForeniscs++) and Celeb-DF~\cite{Celeb_DF_cvpr20}, perform on the same videos and categories of deepfakes that we used in our subjective evaluation. This comparison provides a scientific insight on the differences between human and machine perception of deepfake videos.

To allow researchers to verify, reproduce, and extend our work, we provide the pre-trained models, subjective scores, and the scripts used to analyze the data as an open source package\footnote{Source code: 
\url{https://gitlab.idiap.ch/bob/bob.paper.wifs2020}}.

This paper has the following main contributions:
\begin{itemize}
\item {A comprehensive subjective evaluation and the analysis of human perception of different types of deepfake videos;}
\item {Assessment of Xception and EfficientNet based models on the same videos to compare their performance with human subjects;}
\item {Models, subjective data, and analysis scripts are open source;}
\end{itemize}

%% file: sections/dataset.tex
\section{Data and subjective evaluation}
\label{sec:dataset}

Since the resulted videos produced by automated deepfake generation algorithms vary drastically visually, depending on many factors (training data, the quality of the video for manipulation, and the algorithm itself), we cannot label all deepfakes into one visual category. Therefore, we have manually looked through many videos of Facebook database\footnotemark[2] and pre-selected $60$ deepfake videos, split into five categories depending of how clearly fake they look, with the corresponding $60$ original videos (see examples in Figure~\ref{fig:examples}). 

\begin{figure}[tb]
\centering
\includegraphics[width=0.9\columnwidth]{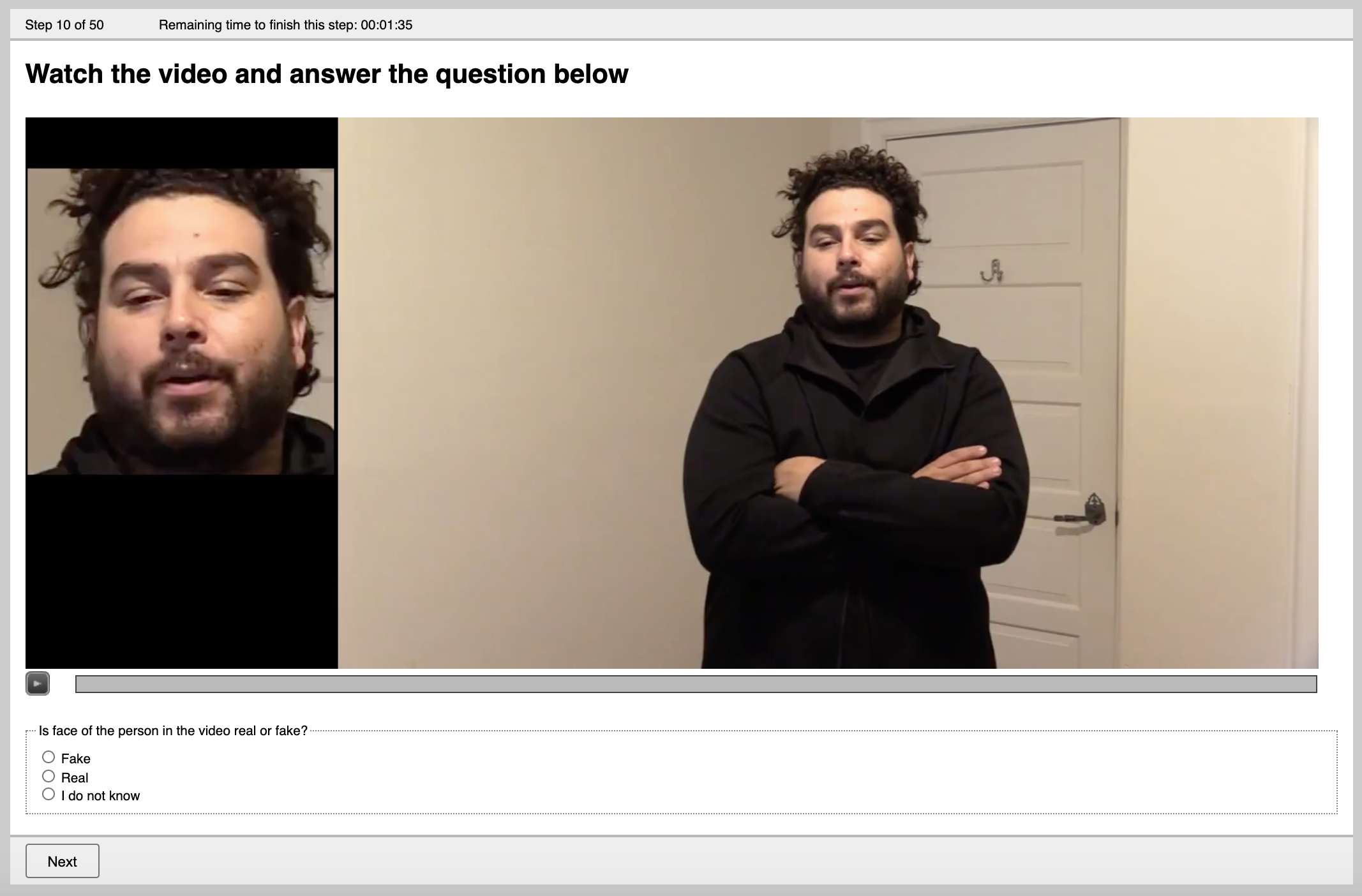}
\caption{Screenshot of one step of subjective evaluation (the video is courtesy of Facebook database).}
\label{fig:screenshot}
\end{figure}

The evaluation was conducted using QualityCrowd 2 framework~\cite{QualityCrowd} designed for crowdsourcing-based evaluations (Figure~\ref{fig:screenshot} shows a screenshot of a typical evaluation step). This framework allows us to make sure subjects watch each video fully at least once and are not able to skip any question. Prior to the evaluation itself, a display brightness test was performed using a method similar to that described in~\cite{Hossfeld2014}. Since deepfake detection algorithms typically evaluate only the face regions cropped using a face detector, to have a comparable scenario, we have also shown to the human subjects cropped face regions next to the original video (see Figure~\ref{fig:screenshot}).

Each of the $60$ na\"ive subjects who participated in the evaluation had to answer the question after watching a given video: ``Is face of the person in the video real or fake?'' with the following options: ``Fake'', ``real'', and ``I do not know.'' Prior to the evaluation, the explanation of the test was given to the subjects with several test video examples of different fake categories and real videos. The $120$ were also split in random batches of $40$ each to reduce the total evaluation time for one subject, so the average time per one evaluation was about $16$ minutes, which is consistent with the standard recommendations.

Due to privacy concerns, we did not collect any personal information from our subjects such as age or gender. Also, the licensing conditions of Facebook database\footnotemark[2] restricted the evaluation to the premises of Idiap research institute, which signed the license agreement not do distribute data outside. Therefore, the subjects consisted of PhD students, scientists, administration, and management of Idiap. Hence the age can be estimated to be between $20$ and $65$ years old and the gender distribution to be of a typical scientific community.

Unlike laboratory-based subjective experiments where all subjects can be observed by operators and its test environment can be controlled, the major shortcoming of the crowdsourcing-based subjective experiments is the inability to supervise participants behavior and to restrict their test conditions. When using crowdsourcing for evaluation, there is a risk of including untrusted data into analysis due to the wrong test conditions or unreliable behavior of some subjects who try to submit low quality work in order to reduce their effort. For this reason, unreliable workers detection is an inevitable process in crowdsourcing-based subjective experiments. There are several methods for identifying the `trustworthiness' of the subject but since our evaluation was conducted within premises of a scientific institute, we only used so called `honeypot' method~\cite{Hossfeld2014,Korshunov2014} to filter out scores from people who did not pay attention at all. Honeypot is a very easy question that refers to the video the subject just watched in the previous steps, e.g., ``what was visible in the previous video?'' with obvious answers that test if a person even looked at the video. Using this question, we filtered out the scores from $5$ people from our final results, hence we ended up with $18.66$ answers on average for each video, which is the number of subjects commonly considered in subjective evaluations.

%% file: sections/results.tex
\section{Subjective evaluation results}

\begin{figure}[tb]
\centering
\includegraphics[width=0.95\columnwidth]{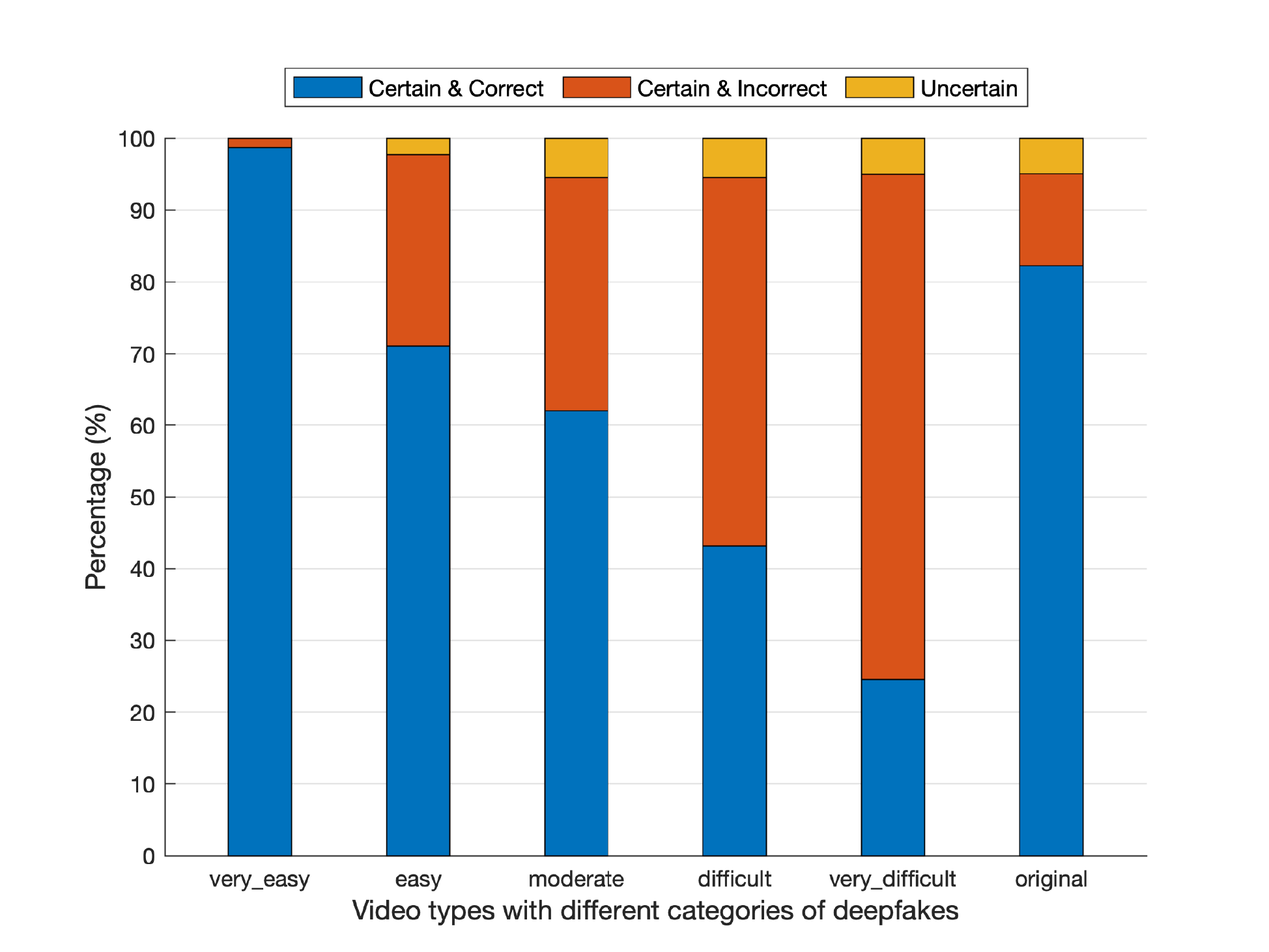}
\caption{Subjective answers for each category of deepfakes and original unaltered videos.}
\label{fig:result_bars}
\end{figure}

For each deepfake or original video, we computed the percentage of answers that were `certain \& correct', when people selected `Real' for an original or `Fake' for a deepfake, `certain \& incorrect' (selected `Real' for a deepfake or `Fake' for an original) and `uncertain', when the selection was `I do not know'. We have averaged those percentages across videos in each category to obtain the final percentages, which are shown in Figure~\ref{fig:result_bars}. From the figure, we can note that the pre-selected deepfake categories, on average, reflect the difficulty level of recognizing them. The interesting results is the low number of uncertain answers, which means people tend to be sure when it comes to judging the realism of a video. And it also means people can be easily spoofed by a good quality deepfake video, since only in $24.5$\% cases `well done' deepfake videos are perceived as fakes, even though these subjects already knew they are looking for fakes. In the scenario, when such deepfake would be distributed to an unsuspected audience (e.g., via social media), we can expect the number of people noticing it to be significantly lower. Also, it is interesting to note that even videos from `easy' category were not as easy to spot ($71.1$\% correct answers) compared to the original videos ($82.2$\%). Overall, we can see that people are better at recognizing very obvious examples of deepfakes or real unaltered videos.

\begin{figure}[tb]
\centering
\includegraphics[width=0.95\columnwidth]{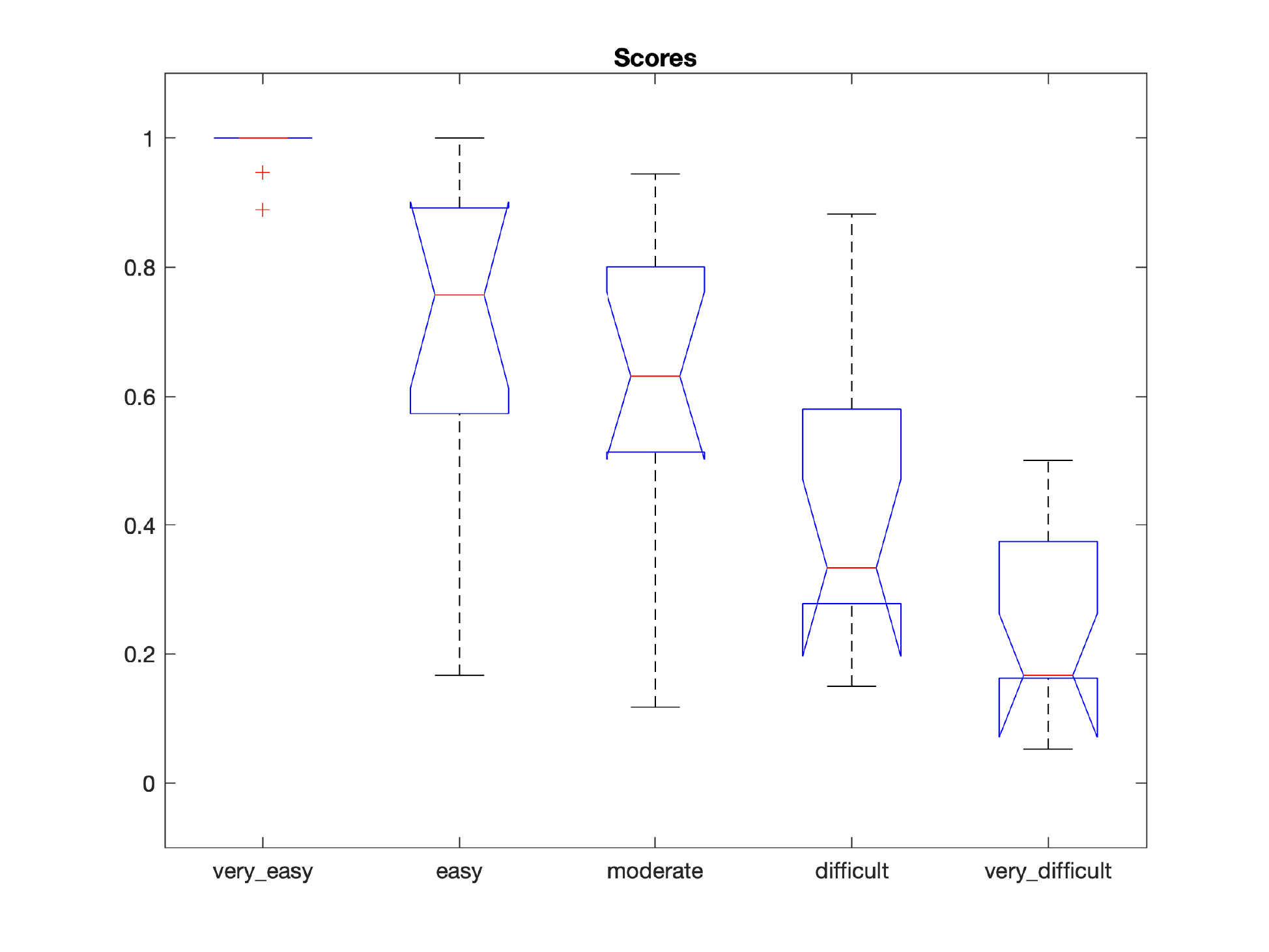}
\caption{Median values with error bars from the ANOVA test performed on subjective scores from five deepfake categories.}
\label{fig:anova}
\vspace{-0.4cm}
\end{figure}

\begin{figure}[tb]
\centering
\includegraphics[width=0.95\columnwidth]{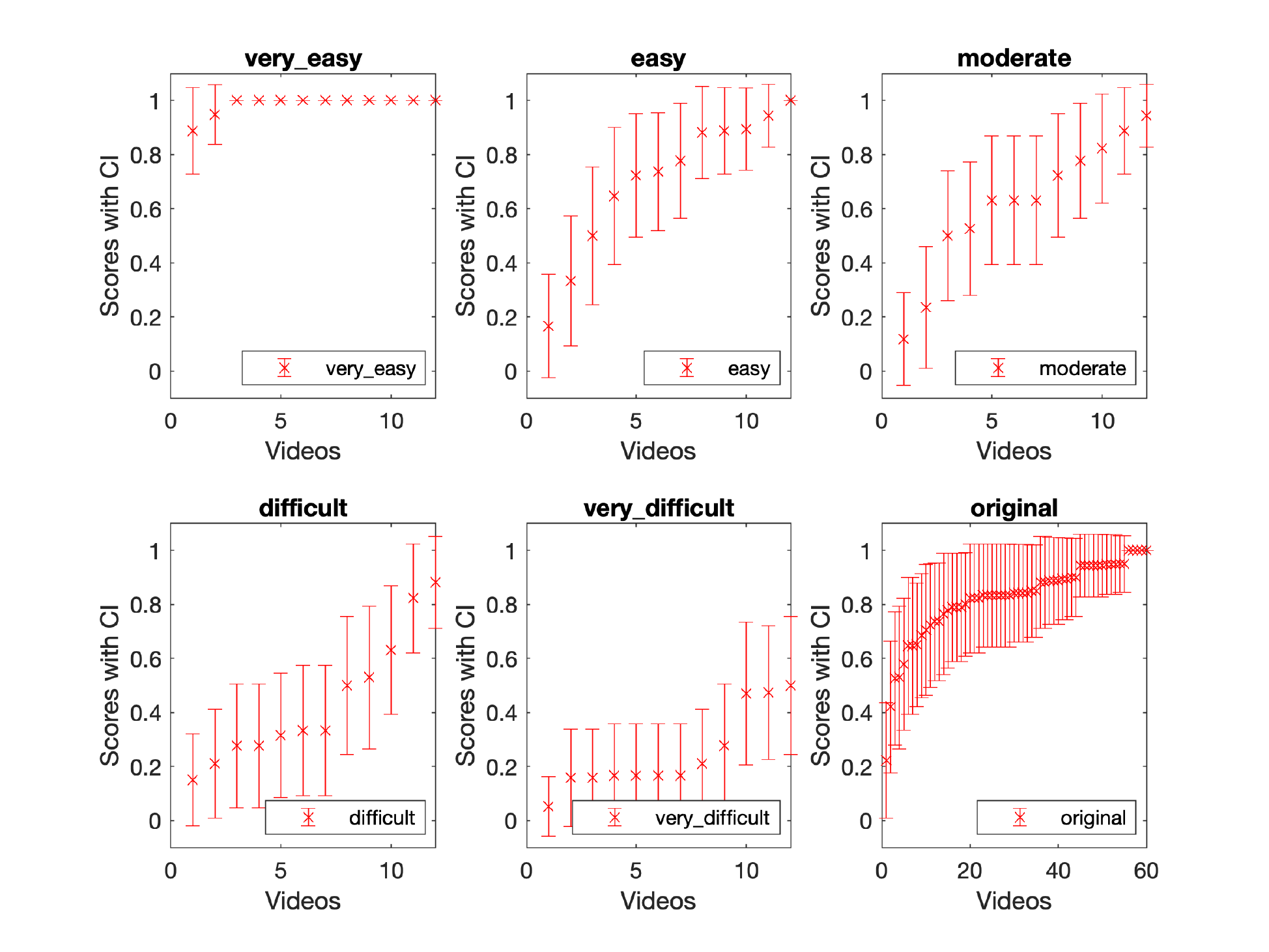}
\caption{Average scores with confidence intervals for each video in every video category.}
\label{fig:errorbars}
\end{figure}

To check whether the difference between videos from the five deepfake categories is statistically significant based on the subjective scores, we performed ANOVA test  with the corresponding box plot shown in Figure~\ref{fig:anova}. The scores were computed for each video (and per category when applicable) by averaging the answers from all corresponding observers. For each correct answer the score is $1$ and for both wrong and uncertain answer the score is $0$.  Please note that the red lines in Figure~\ref{fig:anova} correspond to median values, not average, which what we plotted in Figure~\ref{fig:result_bars}. The \emph{p-value} of ANOVA test is below $4.7e-11$, which means the deepfake categories are significantly different on average. However Figure~\ref{fig:anova} shows that `easy', `moderate', and `difficult' categories have large  scores variations and overlap, which means some of the videos from these categories are perceived similarly. It means some of the deepfake videos could be moved to another category.  This observation is also supported by the Figure~\ref{fig:errorbars} which plots the average scores with confidence intervals (computed using Student's t-distribution~\cite{Hanhart2013}) for each video in the deepfake category ($12$ videos each) and originals ($60$ videos).

\section{Evaluation of algorithms}

For the example of machine vision, we took two state of the art algorithms: based on Xception model~\cite{Chollet2017} and EfficientNet variant B4~\cite{Delp2020} shown to be performing very well on different deepfake datasets and benchmarks~\cite{Roessler2019}. We pre-trained these models for $20$ epochs each on the Google's subset from FaceForensics++ database~\cite{Roessler2019} and Celeb-Df~\cite{Celeb_DF_cvpr20} to demonstrate the impact of different training conditions on the evaluation results. If evaluated on the test sets of the same databases they were trained on, both Xception and EfficientNet classifiers demonstrate a great performance as shown in Table~\ref{tab:algorithms}. We can see that the area under the curve (AUC), which is the common metric used to compare the performance of deepfake detection algorithms, is almost at 100\% in all cases. 

\begin{table}[tb]
\footnotesize
\caption{Area under the curve (AUC) value on the test sets of Google and Celeb-DF databases of Xception and EfficientNet models.}
\label{tab:algorithms}
\centering

\setlength\tabcolsep{1.5pt}
\def\arraystretch{1.3}%

\begin{tabular}{l|l|c|}
\toprule

{\bf Model} & {\bf Trained on} & {\bf AUC (\%) on Test set} \\ \midrule

Xception & Google database & 100.00 \\
Xception & Celeb-DF database & 100.0 \\
EfficientNet & Google database & 99.99 \\
EfficientNet & Celeb-DF database & 100.0 \\
\bottomrule
\end{tabular}
\vspace{-0.5cm}
\end{table}

\begin{figure*}[tbh]
\centering

\subfloat[EfficientNet trained on Google]{\includegraphics[width=0.4\textwidth]{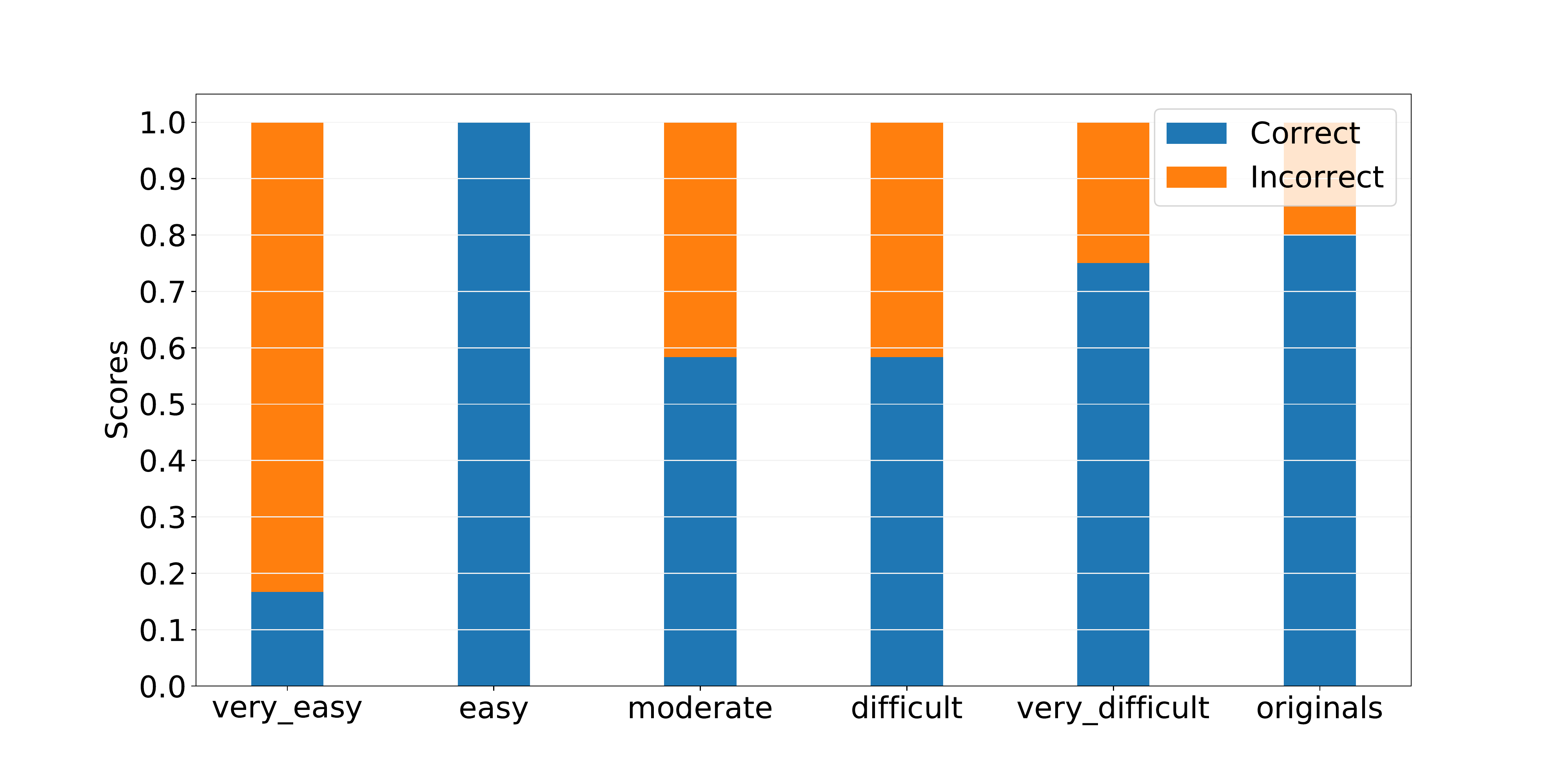}}
\subfloat[EfficientNet trained on Celeb-DF]{\includegraphics[width=0.4\textwidth]{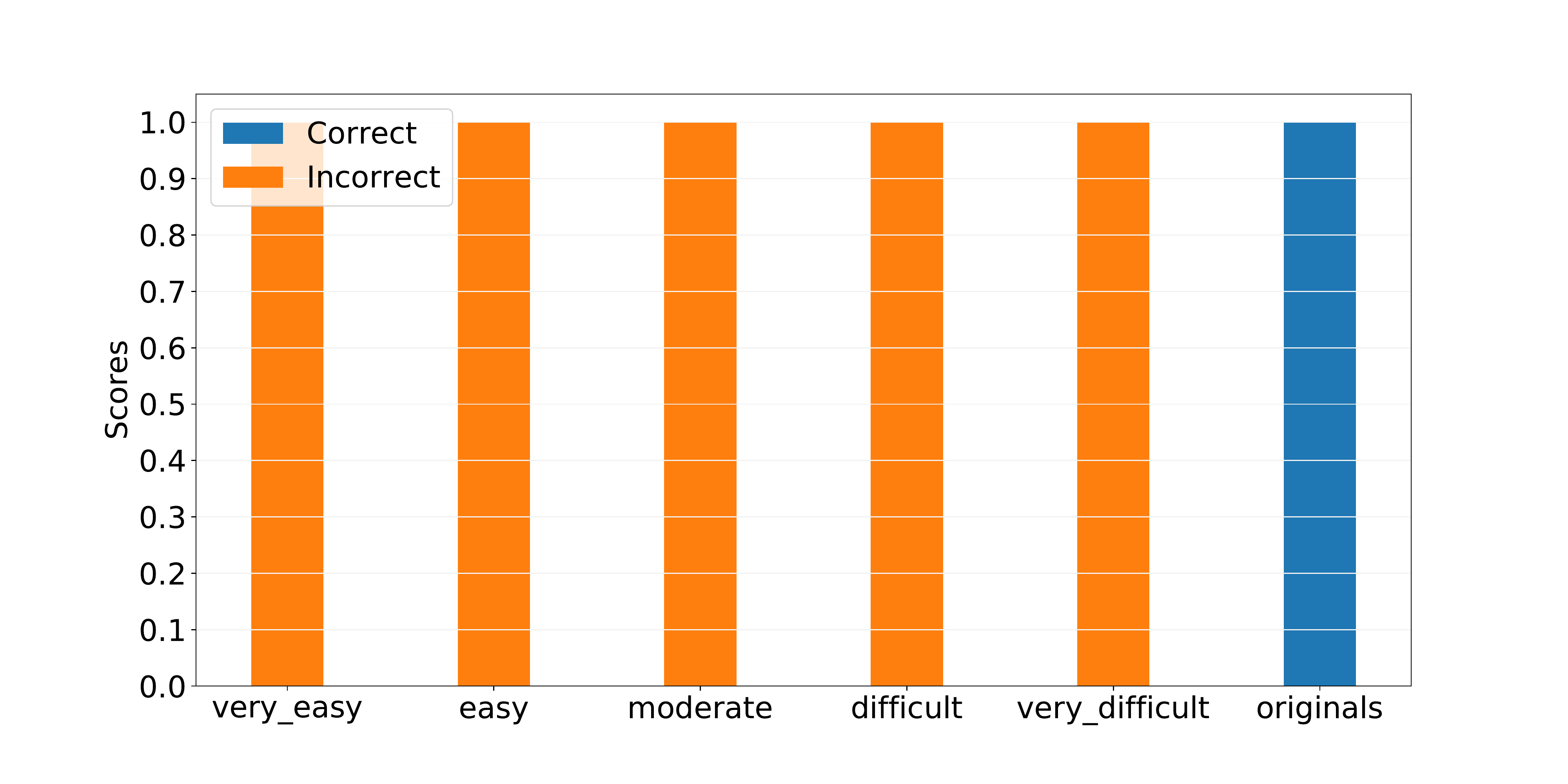}}\\
\subfloat[Xception trained on Google]{\includegraphics[width=0.4\textwidth]{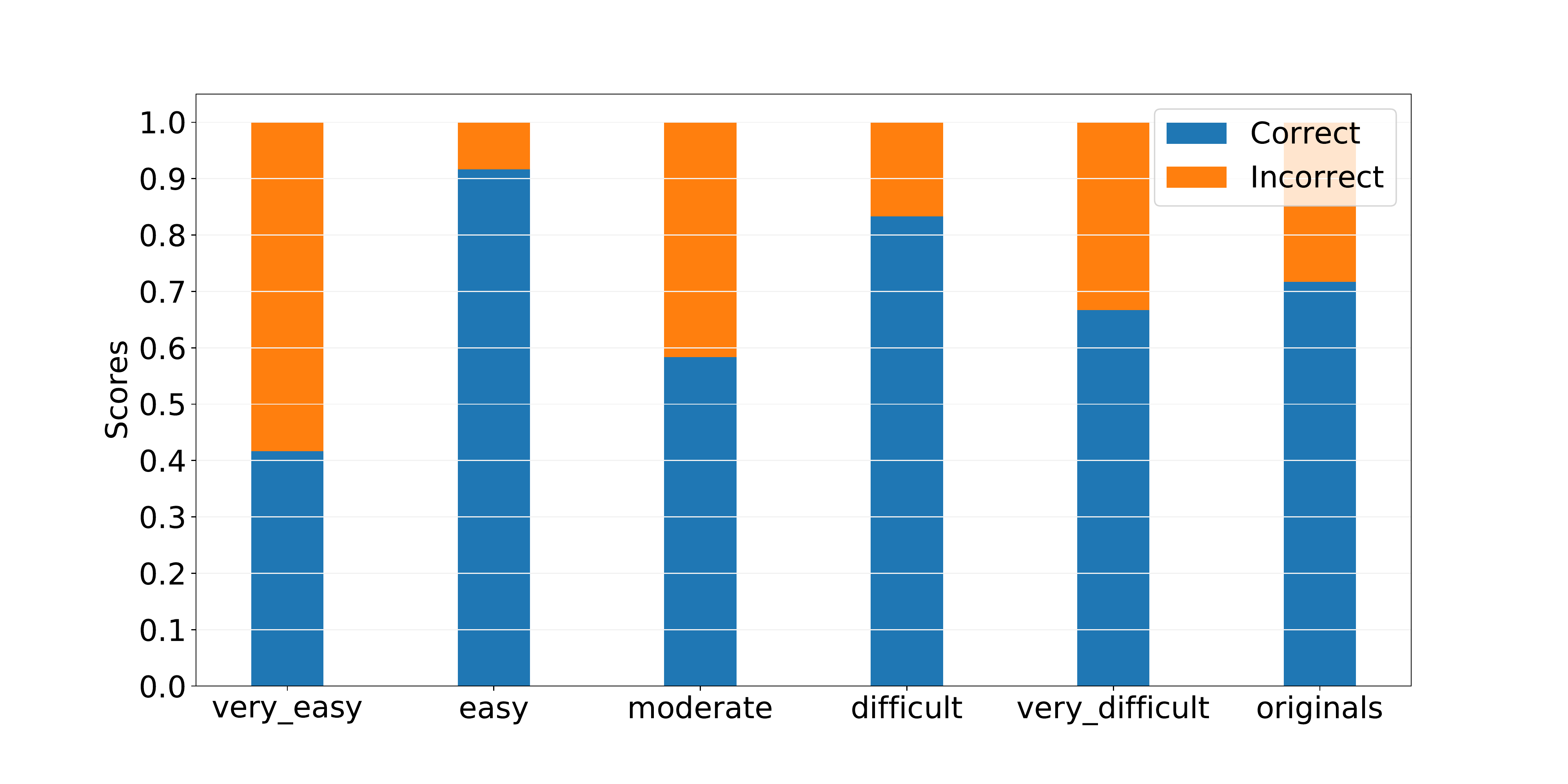}}
\subfloat[Xception trained on Celeb-DF]{\includegraphics[width=0.4\textwidth]{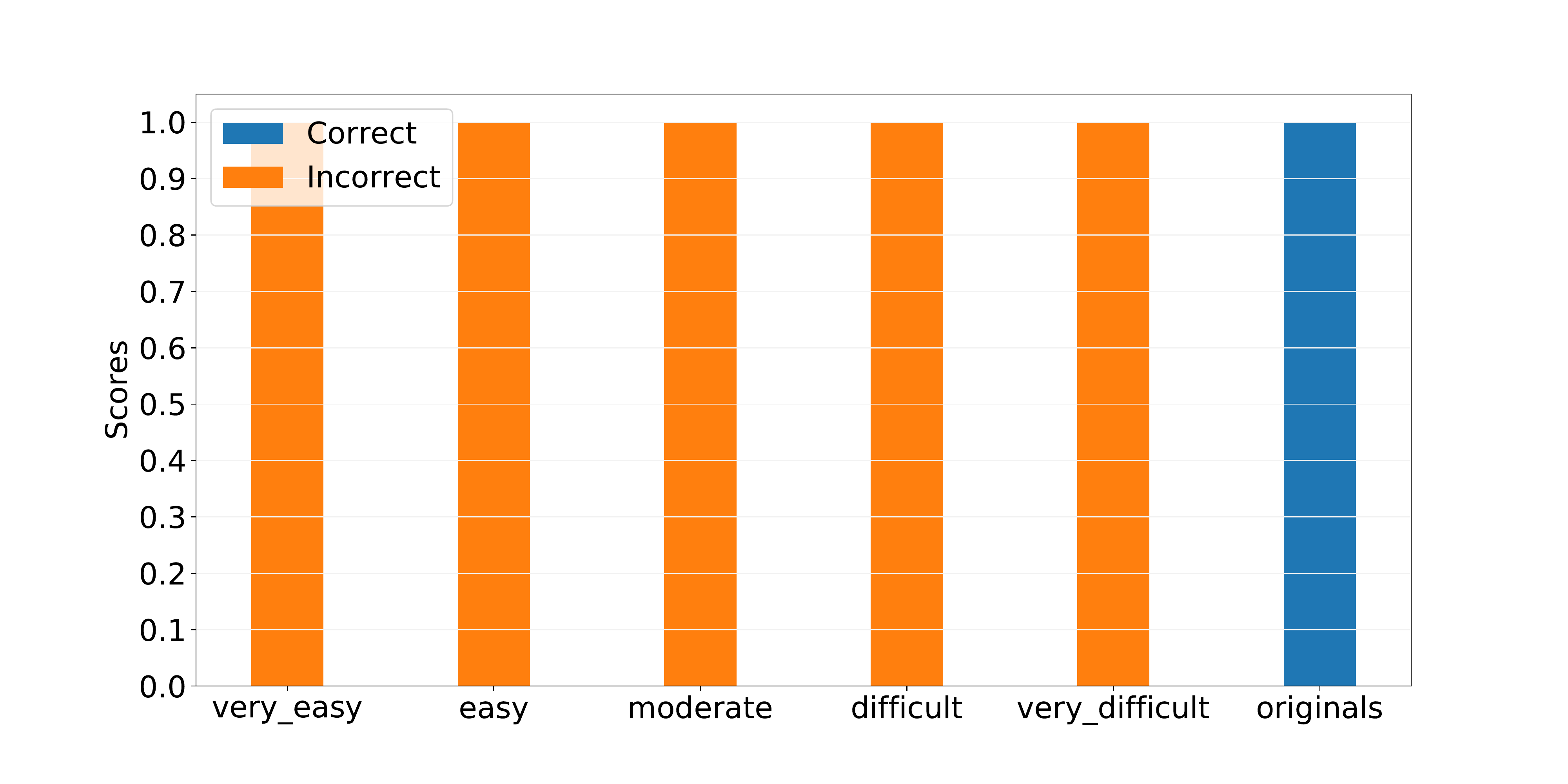}}

\caption{The detection accuracy (the threshold corresponds to FAR $10$\% on development set of the respective database) for each video category from subjective test by Xception and Efficient models pre-trained on Google and Celeb-DF databases.}
\label{fig:algo_results}
\end{figure*}

We evaluated these models on the $120$ videos we used in the subjective test. Since these videos come from Facebook database, they can be considered as unseen data, which is still an obstacle for many DNN classifiers, as they do not generalize well on the unseen data the fact also highlighted in the recent Facebook Deepfake Detection Challenge~\cite{tolosana2020deepfakes}. To compute performance accuracy, we need to select threshold. We chose the threshold corresponding to the false accept rate (FAR) of $10$\%, selected on the development set of the respective database. We selected threshold based on FAR value as oppose to equal error rate (EER) commonly used in biometrics, because many practical deepfake detection or anti-spoofing systems have a low bound requirement on FAR value. In our case, FAR of $10$\% is quite generous. 

Figure~\ref{fig:algo_results} demonstrate the evaluation results of pre-trained Xception and EfficientNet models on the videos from the subject test averaged for each deepfake category and originals (when using threshold corresponding to FAR$=10$\%). In the figure, blue bar corresponds to the percent of correctly detected videos in the given category, and the orange bar correspond to the percent of incorrectly detected. The results for algorithms are very different from the results of the subjective test (see Figure~\ref{fig:result_bars} for the evaluation results by human subjects). The accuracy of the algorithms have no correlation to the visual appearance of deepfakes. The algorithms `see' these videos very differently from how humans perceive the same videos. To a human observer the result may even appear random. We can even notice that all algorithms struggle the most with the deepfake videos that were easy for human subjects. It is evident that the choice of threshold and the training data have major impact on the evaluation accuracy. However, when selecting a deepfake detection system to use in practical scenario, one cannot assume an algorithm's perception will have any relation to the way we think the videos look like. 

\begin{figure*}[tb]
\centering
\includegraphics[width=0.9\textwidth]{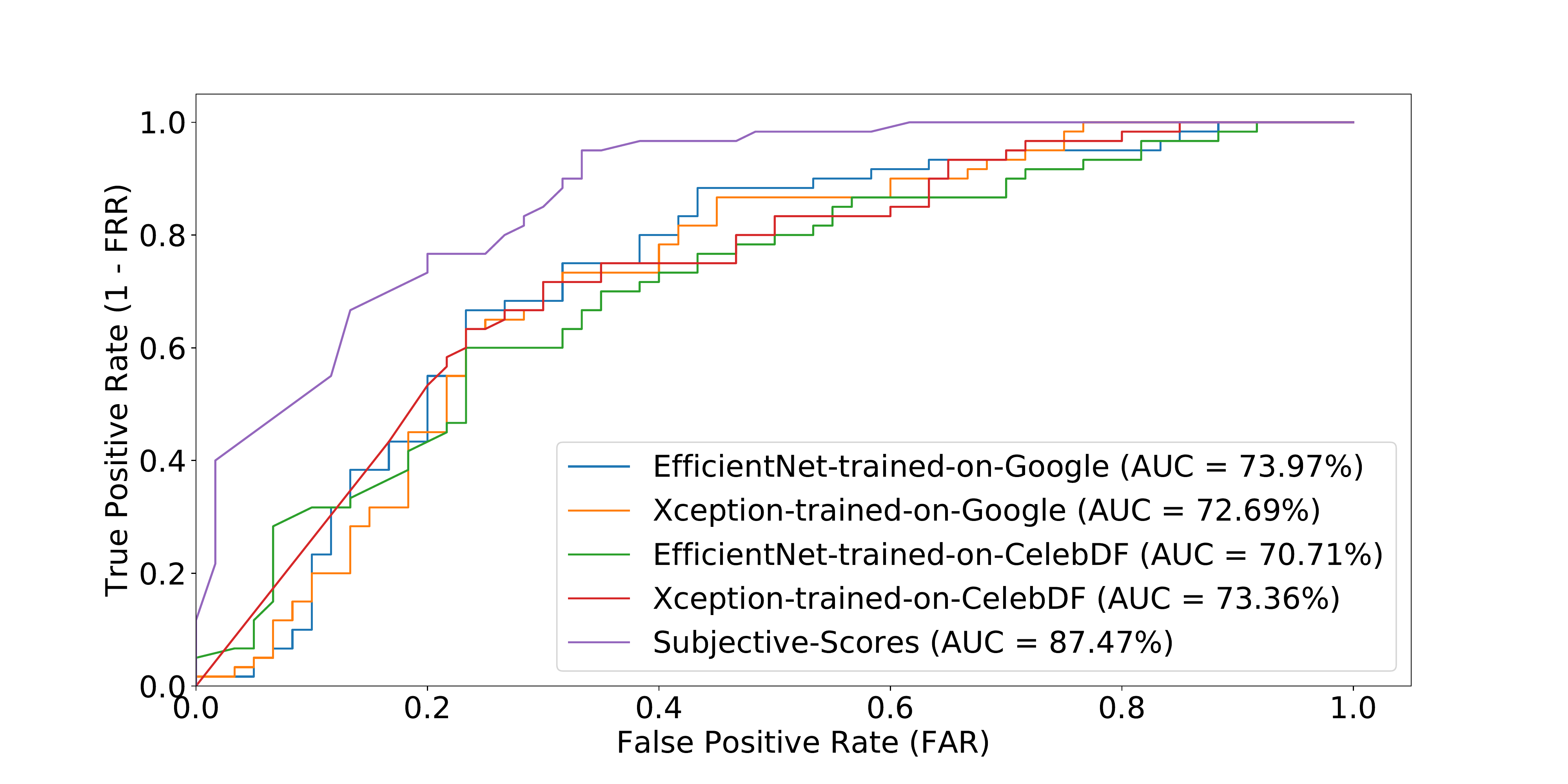}
\caption{ROC curves with the corresponding AUC value of Xception and Efficient models pre-trained on Google and Celeb-DF databases evaluated on all the videos from subjective test.}
\label{fig:roc}
\end{figure*}

If we remove the choice of the threshold and the pre-selected video categories and simply evaluate the models on the $120$ videos from the subjective tests, the receiver operating characteristic (ROC) curve and the corresponding AUC values presented in Figure~\ref{fig:roc}. From this figure, we can note that ROC curves looks `normal', as typical curves for  classifiers that do not generalize well on unseen data, especially taking into account excellent performance on the test sets shown in Table~\ref{tab:algorithms}. Figure~\ref{fig:roc} also shows that human subjects were more accurate at assessing this set of videos since the corresponding ROC curve is consistently higher with the highest AUC value of $87.47$\%.

%% file: sections/conclusion.tex
\section{Conclusion}
In this paper, we presented the results of subjective evaluation of different categories of deepfake videos, ranging from obviously fake to easy being confused with real videos. The videos were manually pre-selected from Facebook database and evaluated by $60$ human subjects. The same videos were also used in the evaluation of two state of the art deepfake detection algorithms based on Xception and EfficientNet models, which were separately pre-trained on Google and Celeb-DF deepfake databases. 

The subjective evaluation demonstrated that people are consistent in the way the perceive different types of deepfakes. Also, the results show that people are confused by good quality deepfakes in $75.5$\% of cases. On the other hand, the algorithms have a totally different perception of deepfakes compared to human subjects. The algorithms struggle to detect many deepfakes, which look obviously fake to humans, while some of the algorithms (depending on the training data and the selected threshold) can accurately detect videos that are difficult for human subjects.

This paper shows that the deepfake generation is already at the level of realism that would confuse the majority of the public, especially in the browser-based viewing scenario. The paper also shows that is important to clearly understand how a given algorithms evaluates data and what conditions impact it performance and in which way. What is even more important is to not confuse and to not anthropomorphize machine vision with human vision, because they are very different and do not correlate with each other.